\documentclass[10pt,twocolumn,letterpaper]{article}

\usepackage{cvpr}
\usepackage{times}
\usepackage{epsfig}
\usepackage{graphicx}
\usepackage{amsmath}
\usepackage{amssymb}

\usepackage{mathtools}
\usepackage[ruled, linesnumbered]{algorithm2e}

\usepackage{multirow}
\usepackage{subfig}
\graphicspath{ {./images/} }
\usepackage{float}
\usepackage{siunitx}
\usepackage{fixltx2e}

\usepackage[pagebackref=true,breaklinks=true,letterpaper=true,colorlinks,bookmarks=false]{hyperref}

\cvprfinalcopy 

\def\cvprPaperID{6263} 

\ifcvprfinal\fi
\begin{document}

\title{Data-Free Adversarial Distillation}

\author{Gongfan Fang\textsuperscript{1}, Jie Song\textsuperscript{1}, Chengchao Shen\textsuperscript{1}, Xinchao Wang\textsuperscript{2}, Da Chen\textsuperscript{3}, Mingli Song\textsuperscript{1}\\
\textsuperscript{1}College of Computer Science and Technology, Zhejiang University, Hangzhou, China\\
\textsuperscript{2}Department of Computer Science, Stevens Institute of Technology, New Jersey, United States \\
\textsuperscript{3}Alibaba Group, Hangzhou, China\\
\{fgf, sjie, chengchaoshen, brooksong\}@zju.edu.cn,\\ xinchao.w@gmail.com, chen.cd@alibaba-inc.com}

\maketitle

\begin{abstract}

Knowledge Distillation (KD) has made remarkable progress in the last few years and become a popular paradigm for model compression and knowledge transfer. However, almost all existing KD algorithms are data-driven, i.e., relying on a large amount of original training data or alternative data, which is usually unavailable in real-world scenarios. In this paper, we devote ourselves to this challenging problem and propose a novel adversarial distillation mechanism to craft a compact student model without any real-world data. We introduce a model discrepancy to quantificationally measure the difference between student and teacher models and construct an optimizable upper bound. In our work, the student and the teacher jointly act the role of the discriminator to reduce this discrepancy, when a generator adversarially produces some ``hard samples'' to enlarge it. Extensive experiments demonstrate that the proposed data-free method yields comparable performance to existing data-driven methods. More strikingly, our approach can be directly extended to semantic segmentation, which is more complicated than classification and our approach achieves state-of-the-art results. The code will be released.


\end{abstract}
\section{Introduction}

Deep learning has made unprecedented advances in a wide range of applications~\cite{krizhevsky2012imagenet,radford2015unsupervised, eigen2014depth, zhu2017unpaired} in recent years. Such great achievement largely attributes to several essential factors, including the availability of massive data, the rapid development of the computing hardware, and the more efficient optimization algorithms. Owing to the tremendous success of deep learning and the open-source spirit encouraged by the research fields, an enormous amount of pretrained deep networks can be obtained freely from the Internet nowadays.

\begin{figure}[h]
 \centering
 \includegraphics[width=8cm]{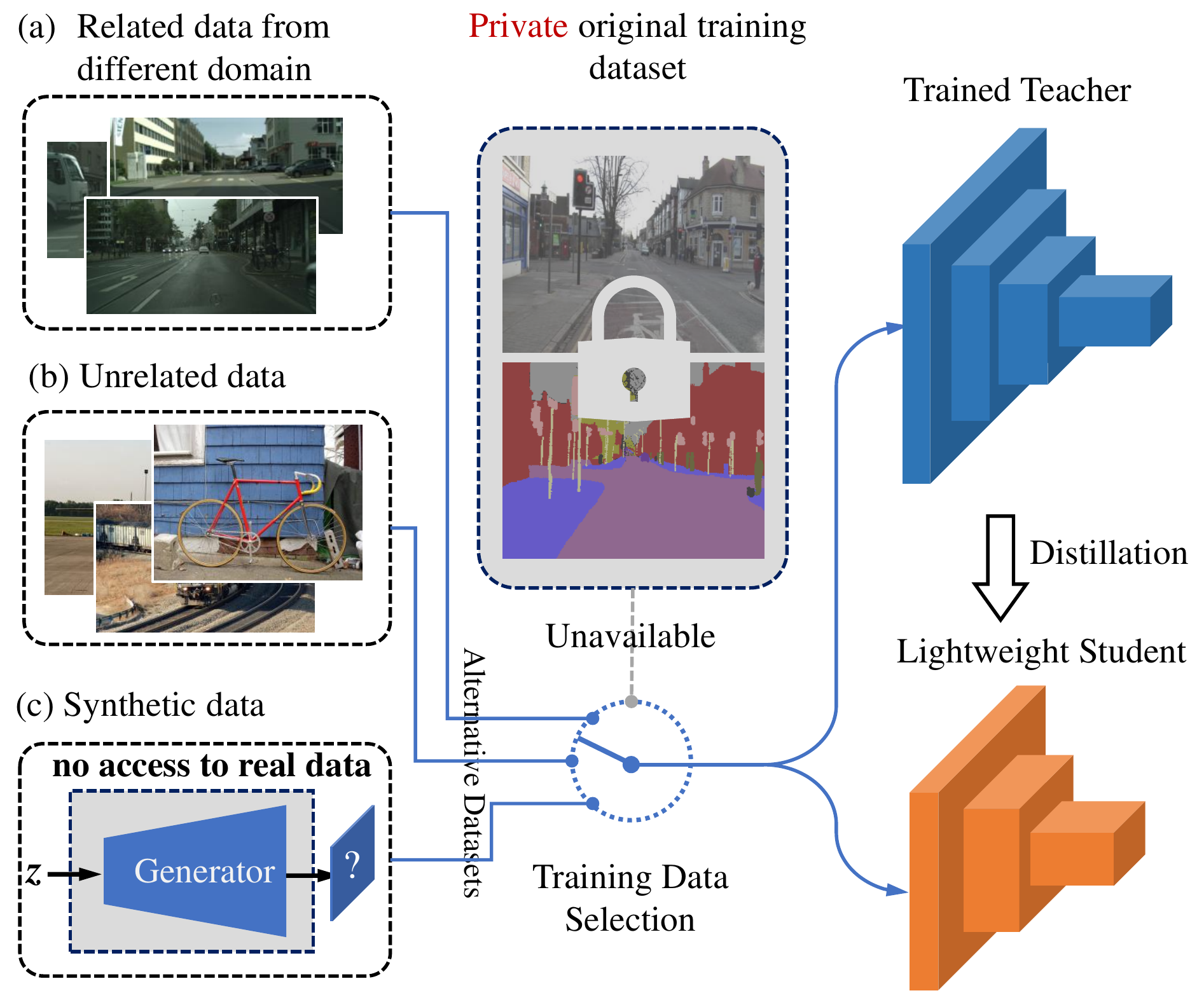}
 \caption{The original training data for pretrained models is usually unavailable to users. In this case, alternative data or synthetic data is used for model compression.}
 \label{fig:introduction}
 \vspace{-4mm}
\end{figure}

However, many problems may occur when we deploy these pretrained models into real-world scenarios. One prominent obstacle is that the pretrained deep model obtained online is usually large in volume, consuming expensive computing resources that we can not afford with the low-capacity edge devices. A large literature has been devoted to compressing the cumbersome deep models into a more lightweight one, from which Knowledge Distillation (KD)~\cite{hinton2015distilling} is one of the most popular paradigms. In most existing KD methods, given the original training data or alternative data similar to the original one, a lightweight student model learns from the pretrained teacher by directly imitating its output. We term these methods data-driven KD. 

Unfortunately, the training data of released pretrained models are often unavailable due to privacy, transmission, or legal issues, as seen in Figure \ref{fig:introduction}. One strategy to deal with this problem is to use some alternative data~\cite{bucilua2006model}, but it leads to a new problem where users are utterly ignorant of the data domain, making it almost impossible to collect similar data. Meanwhile, even if the domain information is known, it is still onerous and expensive to collec a large amount of data. Another compromising strategy in this situation is using somewhat unrelated data for training. However, it drastically deteriorates the performance of the student due to the incurred data bias.

An effective way to avert the problems mentioned above is using synthetic samples, leading to the data-free knowledge distillation~\cite{lopes2017data,chen2019data,nayak2019zero}. Data-free distillation is currently a new research area where traditional generation techniques such as GANS~\cite{goodfellow2014generative} and VAE~\cite{kingma2013auto} can not be directly applied due to the lack of real data. Nayak~\etal~\cite{nayak2019zero} and Chen \etal \cite{chen2019data} have made some pilot studies on this problem. In Nayak's work~\cite{nayak2019zero}, some ``Data Impressions'' are constructed from the teacher model. Besides, in Chen's work~\cite{chen2019data}, they also propose to generate some one-hot samples, which can highly activate the neurons of the teacher model. These exploratory researches achieve impressive results on classification tasks but still have several limitations. For example, their generation constraints are empirically designed based on assumption that an appropriate sample usually has a high degree of confidence in the teacher model. Actually, the model maps the samples from the data space to a very small output space, and a large amount of information is lost. It is difficult to construct samples with a fixed criterion on such a limited space. Besides, these existing data-free methods~\cite{chen2019data,nayak2019zero} only take the fixed teacher model into account, ignoring the information from the student. It means that the generated samples can not be customized with the student model.


To avoid the one-sidedness of empirically designed constraints, we propose a data-free adversarial distillation framework to customize training samples for the student model and teacher model adaptively. In our work, a model discrepancy is introduced to demonstrates the functional difference between models. We construct an optimizable upper bound for the discrepancy so that it can be reduced to train the student model. The contributions of our proposed framework can be summarized as three points:

\begin{itemize}
 \item We propose an adversarial training framework for data-free knowledge distillation. To our knowledge, it is the first approach that can be applied to semantic segmentation.
 \item We introduce a novel method to quantitatively measure the discrepancy between models without any real data.
 \item Extensive experiments demonstrate that the proposed method not only behaves significantly superior to data-free methods, and also yields comparable results to some data-driven approaches.
\end{itemize}

\section{Related Work}
\subsection{Knowledge Distillation (KD)}
Knowledge distillation~\cite{hinton2015distilling} aims at learning a compact and comparable student model from pretrained teacher models. With a teacher-student training schema, it efficiently reduces the complexity and redundancy of the large teacher model. In order to extend the KD framework, researchers have proposed several techniques. According to the requirements of data, we divide those methods into two categories, which are data-driven knowledge distillation and data-free knowledge distillation.

\subsubsection{Data-driven Knowledge Distillation}
Data-driven knowledge distillation requires real data to extract the knowledge from teacher models. Bucilua \etal use a large-scale unlabeled dataset to get pseudo training labels from teacher models~\cite{bucilua2006model}. For generalization, Hinton \etal propose the concept of Knowledge Distillation (KD)~\cite{hinton2015distilling}. In KD, the targets, softened by a temperature $T$, are obtained from the teacher model. The temperature $T$ allows the student model to capture the similarities between different categories.

In order to learn more knowledge, some methods are proposed to utilize intermediate representation as supervision. For example, Romero \etal learn a student model by matching the aligned intermediate representation~\cite{romero2014fitnets}. Moreover, Zagoruyko \etal add a constraint of attention matching to let the student network learn similar attention.~\cite{zagoruyko2016paying}. In addition to classification tasks~\cite{hinton2015distilling, zagoruyko2016paying, furlanello2018born}, knowledge distillation can also be applied to other tasks such as semantic segmentation~\cite{liu2019structured, jiao2019geometry} and depth estimation~\cite{pilzer2019refine}. Recently, it has also been extended to multitasking~\cite{ye2019student, shen2019customizing}. By learning from multiple models, the student model can combine knowledge from different tasks to achieve better performance.

\subsubsection{Data-free Knowledge Distillation}
The data-driven methods mentioned above are difficult to practice if training data is not accessible. Intuitively, the parameters of a model are independent of its training data. It is possible to distill the knowledge out without real data with data-free methods.

To achieve it, Lopes \etal propose to store some metadata during training and reconstruct the training samples during distillation~\cite{lopes2017data}. However, this method still requires metadata during distillation, so it is not completely data-free. Furthermore, Nayak \etal propose to craft Data Impressions (DI) as training data from random noisy images~\cite{nayak2019zero}. They model the softmax space as a Dirichlet distribution and update random noise images to obtain  training data. Another kine of method for data-free distillation is to synthesize training samples with a generator directly. Chen \etal propose DAFL~\cite{chen2019data}, in which the teacher model is fixed as a discriminator~\cite{goodfellow2014generative}. They utilize the generator to construct some training samples, which enable the teacher network to produce highly activated intermediate representations and one-hot predictions. 

\subsection{Generative Adversarial Networks (GANs)}
GANs demonstrate powerful capabilities in image generation~\cite{goodfellow2014generative, radford2015unsupervised, brock2018large} for the past few years. It setups a min-max game between a discriminator and a generator. The discriminator aims to distinguish generated data from real ones when the generator is dedicated to generating more realistic and indistinguishable samples to fool the discriminator. Through the adversarial training, GANs can implicitly measure the difference between two distributions. However, GANs are also facing some problems such as training instability and mode collapse~\cite{arjovsky2017wasserstein, gulrajani2017improved}. Arjovsky \etal propose Wasserstein GAN (WGAN) to make training more stable. WGAN replaces traditional adversarial loss~\cite{goodfellow2014generative} with an approximated Wasserstein distance under 1-Lipschitz constraints so that the gradients of generator will be more stable. Similarly, Qi \etal propose to regularize the adversarial loss with Lipschitz regularization~\cite{qi2017loss}. In practical applications, GANs are highly scalable and can be extended to many tasks such as image-to-image translation~\cite{zhu2017unpaired,isola2017image}, image super-resolution~\cite{wang2018esrgan, ledig2017photo} and domain adaptation~\cite{tzeng2017adversarial, kurmi2019attending}. The powerful capabilities theoretically are qualified for sample generation for data-free knowledge distillation. 

\section{Method}
Harnessing the learned knowledge of a pretrained teacher model $\mathcal{T}(x, \theta^t)$, our goal is to craft a more lightweight student model $\mathcal{S}(x, \theta^s)$ without any access to real-world data. To achieve this, we approximates the model $\mathcal{T}$ with a parameterized $\mathcal{S}$ by minimizing the model discrepancy $\mathcal{D}(\mathcal{T}, \mathcal{S})$, which indicates the differences between the teacher $\mathcal{T}$ and the student $\mathcal{S}$. With the discrepancy, an optimal student model can be expressed as follows:

\begin{equation}
\mathcal{S}^* = \min_\mathcal{S}\mathcal{D}(\mathcal{T}, \mathcal{S})
\label{mae_loss}
\end{equation}

In vanilla data-driven distillation, we design a loss function, e.g., Mean Square Error, and optimize it with real data. The loss function in this procedure can be seen as a specific measurement of the model discrepancy. However, the measurement becomes intractable when the original training data is unavailable. To tackle this problem, we introduce our data-free adversarial distillation (DFAD) framework to approximately estimate the discrepancy so that it can be optimized to achieve data-free distillation. 

\subsection{Discrepancy Estimation}

Given a teacher model $\mathcal{T}(x, \theta^t)$, a student model $\mathcal{S}(x, \theta^s)$ and a specific data distribution $p$, we firstly define a data-driven model discrepancy $\mathcal{D}(\mathcal{T}, \mathcal{S}; p)$:

\vspace{-3mm}
\begin{equation}
 \mathcal{D}(\mathcal{T}, \mathcal{S}; p) = \mathbb{E}_{x\sim p(x)} [ \frac{1}{n} \|\mathcal{T}(x) - \mathcal{S}(x) \|_1 ]
\label{D_TS}
\end{equation}

The constant factor $n$ in Eqn. \ref{D_TS} indicates the number of elements in model output. This discrepancy simply measures the Mean Absolute Error (MAE) of model output across all data points. Note that $\mathcal{S}$ is functionally identical to $\mathcal{T}$ if and only if they produce the same output for any input $x$. Therefore, if $p$ is a uniform distribution $p_u$ covering the whole data space, we can obtain the true model discrepancy $\mathcal{D}^*$. Optimizing such a discrepancy is equivalent to training with random inputs sampled from the whole data space, which is obviously impossible due to the curse of dimensionality. To avert estimating the intractable $\mathcal{D}^*$, we introduce a generator network $\mathcal{G}(z, \theta^g)$ to control the data distribution. Like in GANs, the generator accepts a random variable $z$ from a distribution $p_z$ and generate a fake sample $x$. Then the discrepancy can be evaluated with the generator:

\vspace{-3mm}
\begin{equation}
 \mathcal{D}(\mathcal{T}, \mathcal{S}; \mathcal{G}) = \mathbb{E}_{z\sim p_z(z)} [ \frac{1}{n} \|\mathcal{T}(\mathcal{G}(z)) - \mathcal{S}(\mathcal{G}(z)) \|_1 ]
\label{D_TSG}
\end{equation}

The key idea of our framework is to approximate $\mathcal{D}^*$ with $\mathcal{D}(\mathcal{T}, \mathcal{S}; \mathcal{G})$. In other words, we estimate the true discrepancy between the teacher model and student with a limited number of generated samples. In this work, we divide the generated samples into two types: ``hard sample'' and ``easy sample''. The hard sample is able to produce a relatively larger output differences with the model $\mathcal{T}$ and model $\mathcal{S}$, while the easy sample corresponds to small differences. Suppose that we have a generator $\mathcal{G}^h$ that can always generate hard samples, according to Eqn. \ref{D_TSG}, we can obtain a ``hard sample discrepancy'' $\mathcal{D}^h$. Since hard samples always cause large output differences, it is clear the following inequality is true:

\begin{equation}
 \mathcal{D}^h = \mathcal{D}(\mathcal{T}, \mathcal{S}; \mathcal{G}^h) \ge \mathcal{D}(\mathcal{T}, \mathcal{S}; p_u) = \mathcal{D^*}
\label{upper_bound}
\end{equation}

In this inequality, $p_u$ is the uniform distribution covering the whole data space, which comprises a large amount of hard samples and easy samples. Those easy samples make $\mathcal{D^*}$ numerically lower than $\mathcal{D}^h$ that is estimated on hard samples. The inequality is always established when the generated samples are guaranteed to be hard samples. Under this constant, $\mathcal{D}^h$ provides an upper bound for the real model discrepancy $\mathcal{D}^*$. Note that our goal is to optimize the true model discrepancy $\mathcal{D}^*$, which can be achieved by optimizing its upper bound $\mathcal{D}^h$. 

However, in the process of training the student model $\mathcal{S}$, hard samples will be mastered by the student and converted into easy samples. Hence we need a mechanism to push the generator to continuously generate hard samples, which can be achieved by adversarial distillation. 

\subsection{Advserarial Distillation}
\begin{figure}[t]
 \centering
 \noindent\includegraphics[width=8.2cm]{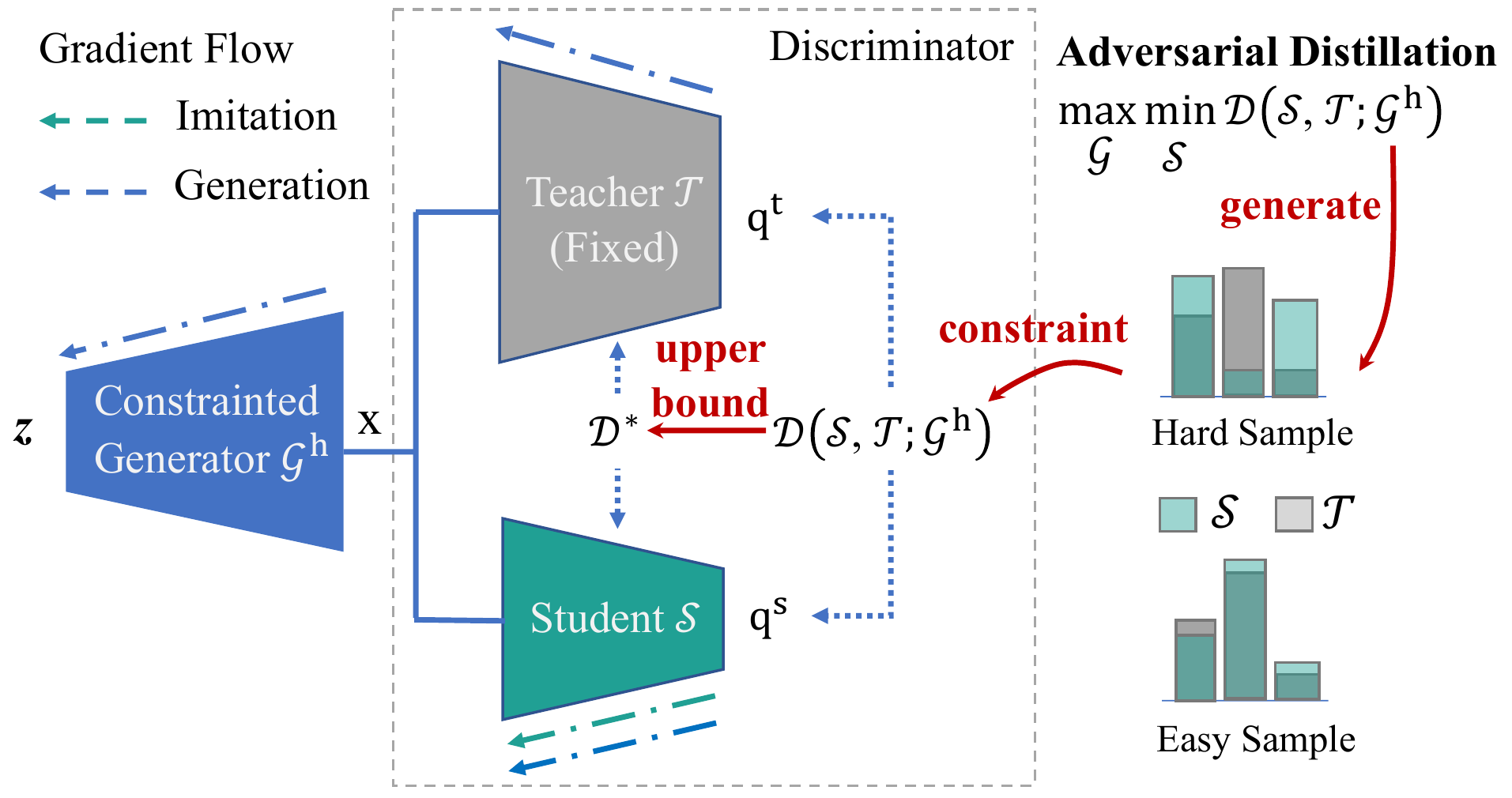}
 \caption{Framework of Data-Free Adversarial Distillation. We construct an upper bound for model discrepancy, under hard sample constraint.}
\label{fig:framework}
\vspace{-3mm}
\end{figure}

In order to maintain the constraints of generating hard samples, we introduce a two-stage adversarial training in this section. Similar to GANs, there is also a generator and a discriminator in our framework. The generator $\mathcal{G}(z, \theta^g)$, as aforementioned, is used to generate hard samples. The student model $\mathcal{S}(x, \theta^s)$, together with the teacher model $\mathcal{T}(x, \theta^t)$ are jointly viewed as the discriminator to measure the hard sample discrepancy $\mathcal{D}(\mathcal{T}, \mathcal{S}; \mathcal{G})$. The adversarial training process consists of two stages: the imitation stage that minimizes the discrepancy and the generation stage that maximize the discrepancy, as shown in Fig. \ref{fig:framework}. 

\subsubsection{Imitation Stage}

In this stage, we fix the generator $\mathcal{G}$ and only update the student $\mathcal{S}$ in the discriminator. we sample a batch of random noises $z$ from Gaussian distribution and construct fake samples $x$ with the generator $\mathcal{G}$. Then each sample $x$ is fed to both the teacher and the student models to produce the output $q^t$ and $q^s$. In classification tasks, $q$ is a vector indicating the scores of different categories. In other tasks such as semantic segmentation, $q$ can be a matrix. 

Actually, there are several ways to define the discrepancy $\mathcal{D}$ to drive the student learning. Hinton \etal utilize the KD loss, which can be Kullback–Leibler Divergence (KLD) or Mean Square Error (MSE), to train the student model. These loss functions are very effective in data-driven KD, yet problematic if directly applied to our framework. An important reason is that, when the student converges on the generated samples, these two loss function will produce decayed gradients, which will deactivate the learning of generator, resulting in a dying minmax game. Hence, the Mean Absolute Error (MAE) between $q^t$ and $q^s$ is used as the loss funtion. Now we can define the loss function for imitation stage as follows:

\vspace{-3mm}
\begin{equation}
 \begin{aligned}
 \mathcal{L}_{IM} & = \mathcal{D}(\mathcal{T}, \mathcal{S}; \mathcal{G}^h) \\
 & = \mathbb{E}_{z\sim p_z(z)} [ \frac{1}{n} \|\mathcal{T}(\mathcal{G}^h(z)) - \mathcal{S}(\mathcal{G}^h(z)) \|_1 ]
 \end{aligned}
\label{im_loss}
\end{equation}

Given the output$q_i^s$ $q_i^t$, The gradient of $|q_i^s - q_i^t|$ with respect to $\theta^s$ is shown in Eqn. \ref{grad_ae}. It simply multiply the gradients with the sign of $q_i^s - q_i^t$  when $q^s$ is very close to $q^t$, which provides stable gradients for the generator so that the vanishing gradients can be alleviated. 

\begin{equation}
 \nabla_{\theta^s} |q_i^s - q_i^t| = sign(q_i^s - q_i^t) \nabla_{\theta^s} q_i^s
 \label{grad_ae}
\end{equation}

Intuitively, this stage is very similar to KD, but the goals are slightly different. In KD, students can greedily learn from the soft targets produced by the teacher, as these targets are obtained from real data~\cite{hinton2015distilling} and contain useful knowledge for the specific task. However, in our setting, we have no access to any real data. The fake samples synthesized by the generator are not guaranteed to be useful, especially at the beginning of training. As aforementioned, the generator is required to produce hard samples to measure the model discrepancy between teacher and student. Another essential purpose of the imitation stage, in addition to learning knowledge from the teacher, is to construct a better search space to force the generator to find new hard samples. 

\subsubsection{Generation Stage}

The goal of the generation stage is to push the generation of hard samples and maintain the constraint for Formula \ref{upper_bound}. In this stage, we fix the discriminator and only update the generator.
It is inspired by the human learning process, where basic knowledge is learned at the beginning, and then more advanced knowledge is mastered by solving more challenging problems. Therefore, in this stage, we encourge the generator to produce more confusing training samples. A straightforward way to achieve this goal is to simply take the negative MAE loss as the objective for optimizing the generator:

\vspace{-3mm}
\begin{equation}
 \begin{aligned}
 \mathcal{L}_{GEN} & = - \mathcal{L}_{IM} \\
 & = - \mathbb{E}_{z\sim p_z(z)} [ \frac{1}{n} \|\mathcal{T}(\mathcal{G}^h(z)) - \mathcal{S}(\mathcal{G}^h(z)) \|_1 ]
 \end{aligned}
\label{gen_loss}
\end{equation}


With the generation loss, the error firstly back-propagates through the discriminator, i.e., teacher and the student model, then the generator, yielding the gradients for optimizing the generator. The gradient from the teacher model is indispensable at the beginning of adversarial training, because the randomly initialized student practically provides no instructive information for exploring hard samples.

However, the training procedure with the objective shown in Eqn.~\ref{gen_loss} may be unstable if the student learning is relatively much slower. By minimizing the objective in Eqn.~\ref{gen_loss}, the generator tends to generate ``abnormal'' training samples, which produce extremely different predictions when fed to the teacher and the student. It deteriorates the adversarial training process and makes the data distribution change drastically. Therefore, it is essential to ensure the generated samples to be normal. To this end, we propose to take the log value of MAE as an adaptive loss function for the generation stage:

\begin{equation}
 \begin{aligned}
 \mathcal{L}_{GEN-ADA} & = -log( \mathcal{L}_{IM} + 1)
 \end{aligned}
\label{ada_gen_loss}
\end{equation}


Different from $\mathcal{L}_{GEN}$ which always encourages the generator to produce hard samples with large discrepancy, in the proposed new objective in Eqn.~\ref{ada_gen_loss}, the gradients of the generator are gradually decayed to zero when discrepancy becomes large. It slow down the training of the generator and make training more stable. Without the log term, we have to carefully adjust the learning rate to make the training as stable as possible.

\subsubsection{Optimization}

\begin{algorithm}[t]
\DontPrintSemicolon
\KwIn{A pretrained teacher model, $\mathcal{T}(x; \theta^t)$}
\KwOut{A comparable student model $\mathcal{S}(x; \theta^s)$}

Randomly initialize a student model $\mathcal{S}(x; \theta^s)$ and a generator $\mathcal{G}(z; \theta^g)$. 

\For{number of training iterations}
{
 \textbf{1. Imitation Stage}

 \For{k steps}
 {
 Generate samples $x$ from $z$ with $\mathcal{G}(z; \theta^g)$;
 
 Calculate model discrepancy with Eqn. \ref{im_loss};
 
 Update $\theta^s$ to \textbf{minimize discrepancy} with
 
 \center $\nabla_{\theta^{s}} \mathcal{D}(\mathcal{T}, \mathcal{S}; \mathcal{G}^h)$
 }

\let\oldnl\nl
\newcommand{\nonl}{\renewcommand{\nl}{\let\nl\oldnl}}
\textbf{2. Generation Stage} 

 Generate samples $x$ from $z$ with $G(z; \theta^g)$; 
 
 Calculate negative discrepancy with Eqn. \ref{gen_loss};
 
 Update $\theta^g$ to \textbf{maximize discrepancy} with 

 \nonl \centering $\nabla_{\theta^{g}} - \mathcal{D}(\mathcal{T}, \mathcal{S}; \mathcal{G}^h)$
}
\caption{Data Free Adversarial Distillation}\label{alg:DFAD}
\end{algorithm}

\noindent \textbf{Two-stage training}. The whole distillation process is summarized in Algorithm \ref{alg:DFAD}. Our framework trains the student and the generator by repeating the two stages. It begins with the imitation stage to minimize $\mathcal{L}_{IM}$. Then in the generation stage, we update the generator to maximize $\mathcal{L}_{GEN}$. Based on the learning progress of student model, the generator crafts hard samples to further estimate the model discrepancy. The competition in this adversarial game drives the generator to discovers missing knowledge, leading to complete knowledge. After several steps of training, the system will ideally reach a balance point, at which the student model has mastered all hard samples, and the generator is not able to differentiate between the two models $\mathcal{S}$ and $\mathcal{T}$. In this case, $\mathcal{S}$ is functionally identical to $\mathcal{T}$.

\noindent \textbf{Training Stability} It is essential to maintain stability in adversarial training.In the imitation stage, we update the student model for $k$ times so as to ensure its convergence. However, since the generating samples are not guaranteed to be useful for our tasks, the value of k cannot be set too large, as it leads to an extraordinarily biased student model. We find that setting k to 5 can make training stable. In addition, we suggest using adaptive loss $\mathcal{L}_{GEN-ADA}$ in dense prediction tasks, such as segmentation, in which each pixel will provide statistical information for adjusting the gradient. In classification tasks, only a few samples are used to calculate the generation loss and the statistical information is not accurate, hence the $\mathcal{L}_{GEN}$ is more prefered.

\noindent \textbf{Sample Diversity} Unlike GANs, our approach naturally maintains diversity of generated samples. When mode collapse occurs, it is easy for students to fit these duplicated samples in our framework, resulting in a very low model discrepancy. In this case, the generator is forced to generate different samples to enlarge the discrepancy. 

\section{Experiments}
We conduct extensive experiments to verify the effectiveness of the proposed method, in which knowledge distillation on two types of models are explored: the classification models and the segmentation model.

\subsection{Experimental Settings}

\begin{table}[t]
 \centering
 \begin{tabular}{ r|c|c|c }
 \hline
 \hline
 \bf Dataset & \bf Model & \bf REL &\bf UNR\\
 \hline
 MNIST & LeNet-5 & SVHN & FMNIST \\
 CIFAR & ResNet & STL10 & Cityscapes \\
 Caltech101 & ResNet & STL10 & Cityscapes \\
 CamVid & DeeplabV3 & Cityscapes & VOC2012 \\
 NYUv2 & DeeplabV3 & SunRGBD & VOC2012 \\
 \hline
 \end{tabular}
 \caption{The datasets and model architectures used in experiments. REL and UNR correspond to related alternative data and unrelated data respectively.}
 \label{dst_info}
 \vspace{-3mm}
\end{table}

\begin{table*}[t]
 \centering
 \begin{tabular}{p{14.5mm}|p{11mm}|p{19mm}||p{11mm}|p{19mm}||p{11mm}|p{19mm}||p{11mm}|p{19mm} }
 \hline
 \hline
 & \multicolumn{2}{c||}{MNIST} & \multicolumn{2}{c||}{CIFAR10} & \multicolumn{2}{c||}{CIFAR100} & \multicolumn{2}{c}{Caltech101} \\
 \hline
 \bf Method & \bf FLOPs & \bf Accuracy & \bf FLOPs & \bf Accuracy & \bf FLOPs & \bf Accuracy & \bf FLOPs & \bf Accuracy \\
 \hline
 Teacher & 433K & 0.989 & 1.16G & 0.955 & 1.16G & 0.775 & 1.20G & 0.766 \\
 KD-ORI & 139K & 0.988 \small $\pm$ 0.001 & 557M & 0.939 \small $\pm$ 0.011 & 558M & 0.733 \small $\pm$ 0.003 & 595M & 0.775 \small $\pm$ 0.002 \\
 KD-REL & 139K & 0.960 \small $\pm$ 0.006 & 557M & 0.912 \small $\pm$ 0.002 & 558M & 0.690 \small $\pm$ 0.004 & 595M & 0.748 \small $\pm$ 0.003 \\
 KD-UNR & 139K & 0.957 \small $\pm$ 0.007 & 557M & 0.445 \small $\pm$ 0.012 & 558M & 0.133 \small $\pm$ 0.003 & 595M & 0.352 \small $\pm$ 0.015 \\
 \hline
 RANDOM & 139K & 0.747 \small $\pm$ 0.033 & 557M & 0.101 \small $\pm$ 0.002 & 558M & 0.015 \small $\pm$ 0.001 & 595M & 0.010 \small $\pm$ 0.000 \\
 DAFL & 139K & 0.981 \small $\pm$ 0.001 & 557M & 0.885 \small $\pm$ 0.003 & 558M & 0.614 \small $\pm$ 0.005 & 595M & FAILED \\
 Ours & 139K & \bf 0.983 \small $\pm$ 0.002 & 557M & \bf 0.933 \small $\pm$ 0.000 & 558M & \bf0.677 \small $\pm$ 0.003 & 595M & \bf 0.735 \small $\pm$ 0.008 \\
 \hline
 \end{tabular}
 \caption{Test accuracy of different distillation methods on several classification datasets. } \label{cls-results}
\end{table*}

\subsubsection{Models and Datasets}
We adopt the following six pretrained models to demonstrate the effectiveness of the proposed method: MNIST~\cite{lecun1998mnist}, CIFAR10~\cite{krizhevsky2009learning}, CIFAR100~\cite{krizhevsky2009learning}, Caltech101 for classification and CamVid~\cite{brostow2008segmentation, brostow2009semantic}, NYUv2~\cite{silberman2012indoor} for semantic segmentation. Here the models are named after the corresponding training data.

\noindent\textbf{MNIST}. MNIST~\cite{lecun1998mnist} is a simple image dataset for recognition of handwritten digits containing 60,000 training images and 10,000 test images from 10 categories. Following \cite{lopes2017data,chen2019data}, we use a LeNet-5 as the pretrained teacher model and use a LeNet-5-Half as the student model. \\
\textbf{CIFAR10} and \textbf{CIFAR100}. CIFAR10~\cite{krizhevsky2009learning} and CIFAR100 both contain 60,000 RGB images. Among them, 50,000 images are used for training and 10,000 for testing. CIFAR10 contains 10 classes when CIFAR100 contains 100 classes. Due to the limitations of the small resolution, we use a modified ResNet-34~\cite{he2016deep} as our teacher, which has only three downsample layers. We utilize a ResNet-18 as our student model.\\
\textbf{Caltech101}. Caltech101~\cite{fei2006one} is a classification dataset. There are 101 categories, each of which contains at least 40 images. We randomly split the dataset into two parts: a training set with 6982 images and a test set with 1695 images. During training, the images are resized and cropped to $128\times128$. We use the standard ResNet-34 architecture as the teacher model and use ResNet-18 as the student model.\\
\textbf{CamVid}. Camvid~\cite{brostow2008segmentation, brostow2009semantic} is a road scene segmentation dataset, consisting of 367 training and 233 testing RGB images. There are 11 categories in CamVid, such as road, cars, poles, traffic lights, etc. The original resolution of images is $720\times960$. Due to the difficulty in generating high-resolution images, we resize the short side to 256 and train our teacher with $128\times128$ random crop. The teacher model is a DeepLabV3~\cite{chen2017rethinking} model with ResNet-50~\cite{he2016deep} as backbone. For student model, we adopt a Mobilenet-V2~\cite{howard2017mobilenets} as the backbone.\\
\textbf{NYUv2}. The NYUv2~\cite{silberman2012indoor} is collected for indoor scene parsing. It provides 1449 labeled RGB-D images with 13 categories and 407024 unlabeled images. We use 795 pixel-wise labeled images to train our teacher and use the left 654 images as the test set. Similar to CamVid, we also resize and crop the images to $128\times128$ blocks for training and use the DeeplabV3 as our model architecture.

\subsubsection{Implementation Details and Evaluation Metrics}

Our method is implemented with Pytorch~\cite{paszke2017automatic} on an NVIDIA Titan Xp. For training, We use SGD with momentum 0.9 and weight decay 5e-4 to update student models. The generator in our method is trained with Adam~\cite{kingma2014adam}. During training, the learning rates of SGD and Adam are decayed by 0.1 for every 100 epochs. In order to measure function discrepancy, we use a large batch size for adversarial training. The batch size is set to 512 for MNIST, 256 for CIFAR, and 64 for other datasets. In our experiments, all models are randomly initialized except that, in semantic segmentation tasks, the backbone of teacher models are pretrained on ImageNet~\cite{krizhevsky2012imagenet}. More detailed hyperparameter settings for different datasets can be found in supplementary materials.

To evaluate our methods, we take the accuracy of prediction as our metric for classification tasks. Furthermore, for semantic segmentation, we calculate Mean Intersection over Union (mIoU) on the whole test set. 

\subsubsection{Baselines}
A bunch of baselines is compared to demonstrate the effectiveness of our proposed method, including both data-driven and data-free methods. The baselines are briefly described as follows.

\noindent\textbf{Teacher}: the given pretrained model which serves as the teacher in the distillation process.\\
\textbf{KD-ORI}: the student trained with the vanilla KD~\cite{hinton2015distilling} method on original training data.\\
\textbf{KD-REL}: the student trained with the vanilla KD on an alternative data set which is similar to the original training data.\\
\textbf{KD-UNR}: the student trained with the vanilla KD on an alternative data set which is unrelated to the original training data.\\
\textbf{RANDOM}: the student trained with randomly generated noise images. \\
\textbf{DAFL}: the student trained with DAta-Free Learning~\cite{chen2019data} without any data without any real data.\\

\subsection{KD in Classification Models}

The testing accuracy of our methods and the compared baselines are provided in Table 2.
In order to eliminate the effects of randomization, we repeat each experiment for 5 times and record average value and standard deviation of the highest accuracy. The first part of the tables gives the results on data-driven distillation methods. KD-ORI requires the original training data when KD-REL and KD-UNR use some unlabeled alternative data for training. In KD-REL, the training data should be similar to the original training data. However, the domain different between alternative and original data is unavoidable, which will result in incomplete knowledge. As shown in the table, the accuracy of KD-REL is slightly lower than KD-ORI. Note that in our experiments, the original training data is available, so that we can easily find some similar data for training. Nevertheless, in the real world, we are ignorant of the domain information, which makes it impossible to collect similar data. In this case, the blindly collected data may contain many unrelated samples, leading to the KD-UNR methods. The incurred data bias makes training very difficult and deteriorates the performance of the student model.

The second part of the table shows the results of data-free distillation methods.
We compare our methods with DALF~\cite{chen2019data} using their released code. In our experiment, we set the batch size to 256 for CIFAR and 64 for Caltech101 train each model for 500 epochs. Our adversarial learning method achieves the highest accuracy among the data-free methods, and the performance is even comparable to those data-driven methods. Note that we set the batch size of Caltech101 to 64, DAFL methods failed in this case when our method is still able to learn a student model from the teacher. The influence of different batch sizes can be found in supplementary materials.


\begin{figure}[t]
 \centering
 \includegraphics[width=8cm]{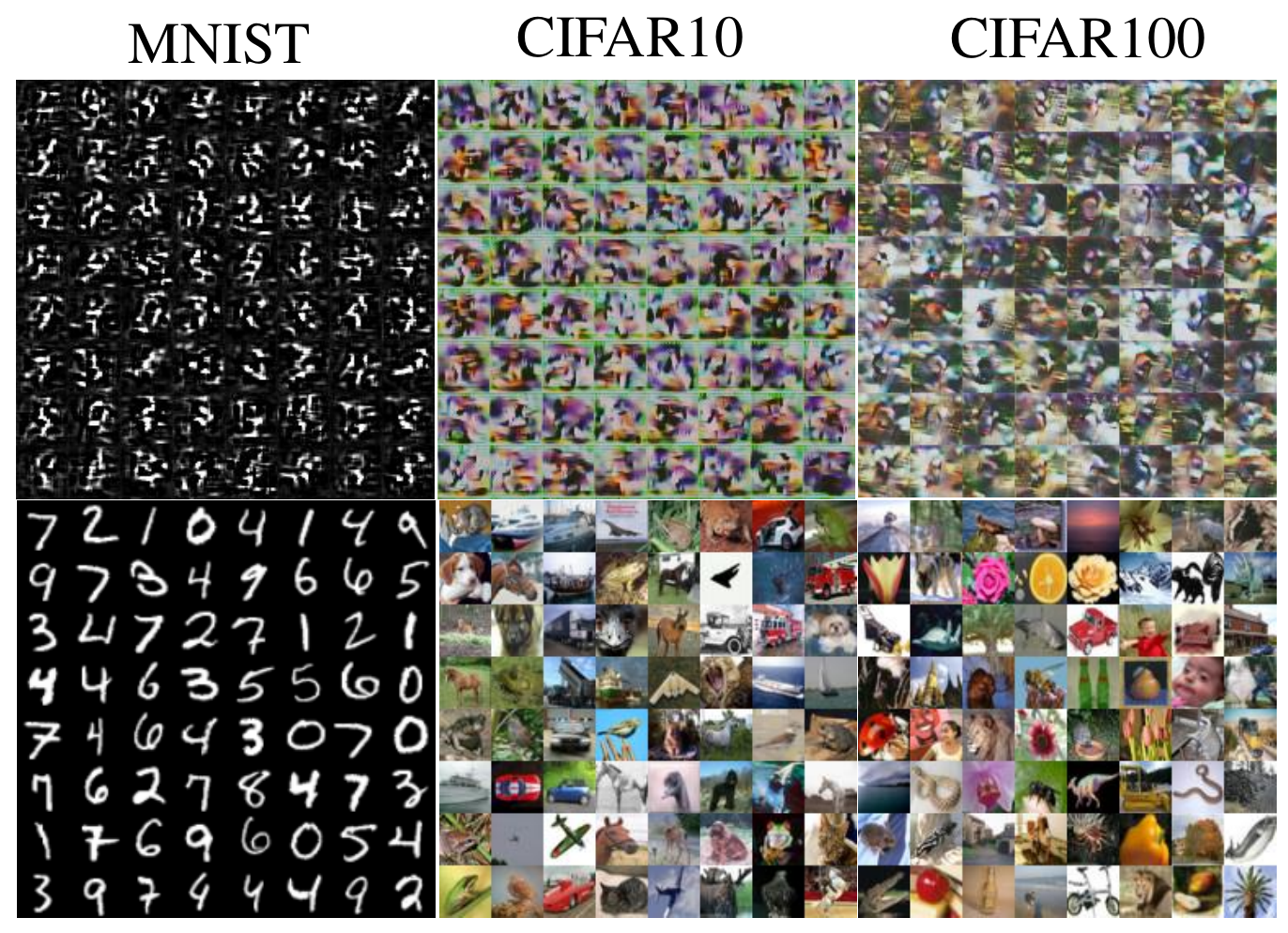}
 \caption{Generated samples on MNIST, CIFAR10 and CIFAR100. The images in the second row are sampled from real data.}
 \label{fig:generated-samples}
 \label{fig:manifold}\vspace{-3mm}
\end{figure}

\noindent\textbf{Visualization of Generated Samples.} The generated samples and real samples are shown in Figure ~\ref{fig:generated-samples}. The images in the first row are produced by the generator during adversarial learning, and the real images are listed in the second row. Although those generated samples are not recognizable by humans, they can be used to craft a comparable student model. It means that using realistic samples are not the only way for knowledge distillation. Comparing the generated samples on CIFAR10 and CIFAR100, we can find that the generator on CIFAR100 produces more complicated samples than on CIFAR10. As the difficulty of classification increases, the teacher model becomes more knowledgeable so that, in adversarial learning, the generator can recover more complicated images. As mentioned above, the diversity of generated samples are guaranteed by the adversarial loss. In our results, the generator does maintain a perfect image diversity, and almost every generated image is different.

\noindent\textbf{Comparison between Loss Functions.} It is essential to keep the balance of the adversarial game. An appropriate adversarial loss should provide stable gradients during training. In this experiment, four candidates are explored, which are MAE, MSE, KLD, and MSE+MAE. By comparing the accuracy curves of the different loss function, we find that MAE indeed provides the best results owing to its stable gradient for generator.

\begin{figure}[t]
 \includegraphics[width=8.2cm]{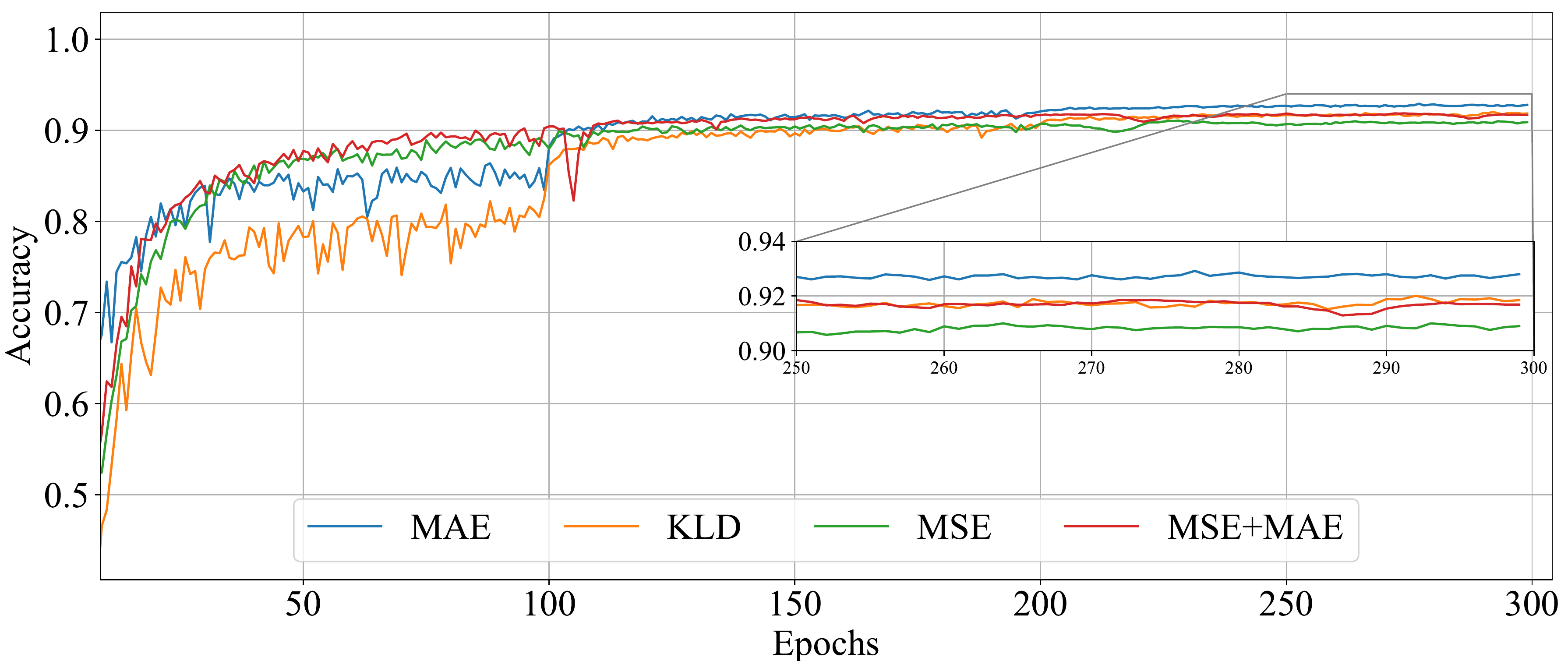}
 \caption{The accuracy curve of different loss functions on CIFAR10. MAE achieves the best performance among those loss candidates. }
 \label{fig:manifold}\vspace{-3mm}
\end{figure}

\subsection{KD in Segmentation Models}

\begin{figure*}[t]
 \centering
 \noindent\includegraphics[width=170mm]{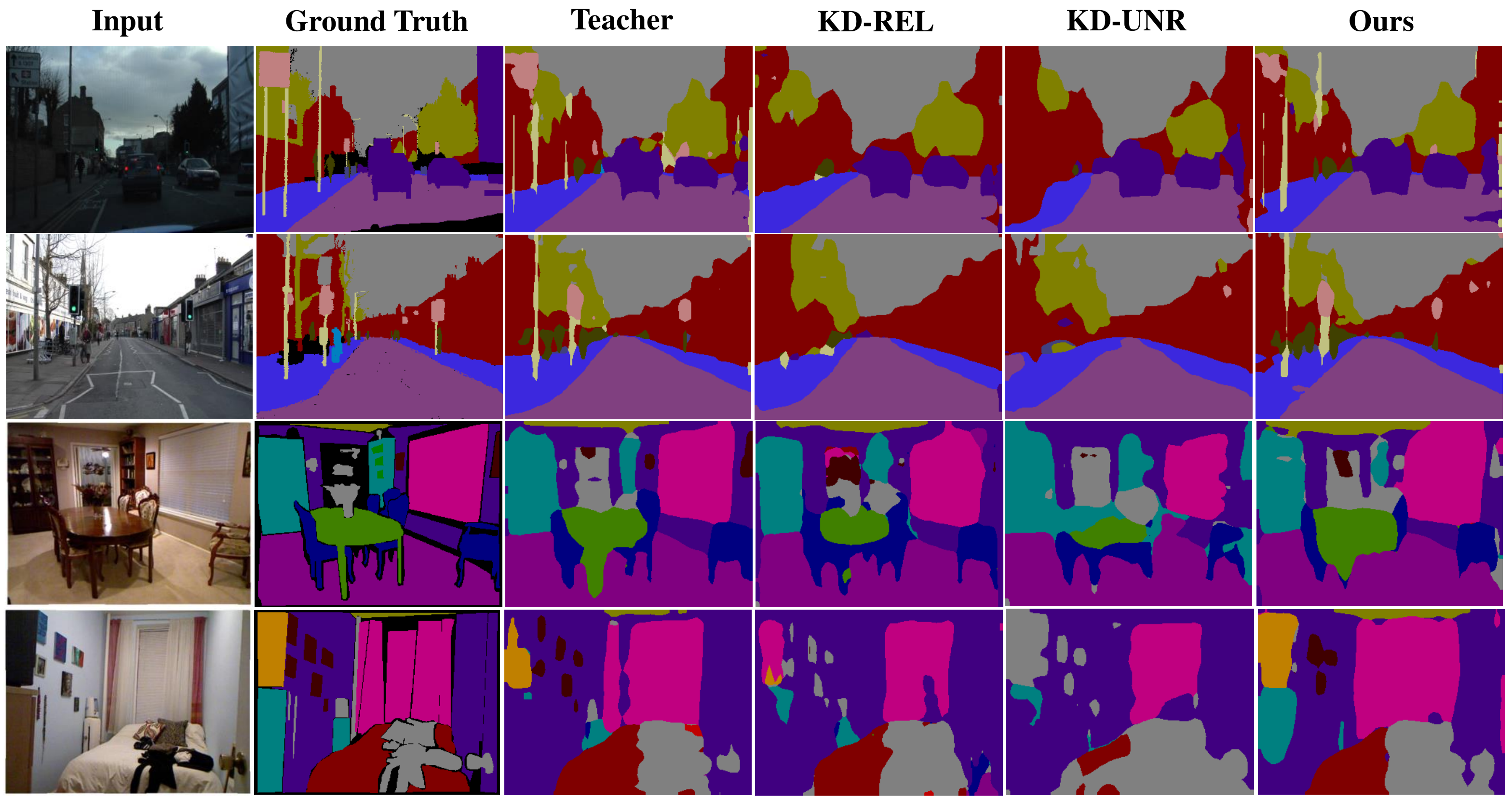}
 \caption{Segmentation results on CamVid and NYUv2. All baseline methods in the figure are data-driven and our framework achieves the best performance when the original training data is not available.}
 \label{fig:seg-visualization}
 \vspace{-3mm}
\end{figure*}

Our method can be naturally extended to semantic segmentation tasks. In this experiment, we adopt ImageNet-pretrained ResNet-50 to initialize the teacher model and train all student models from scratch. All Models in data-driven methods are trained with $128\times128$ cropped images. For data-free methods, $128\times128$ images are directly generated for training. Table \ref{seg-miou} shows the performance of the student model obtained with different methods. We can see that, on CamVid, our method obtains a competitive student model even compared with KD-ORI, which requires the original training data. On NYUv2, our approach goes beyond KD-UNR and all data-free methods, although not comparable to KD-ORI. In fact, our method is the first data-free distillation method proposed to work on segmentation tasks. 

\begin{table}[t]
 \centering
 
 \begin{tabular}{p{14mm}|p{11mm}|p{11mm}||p{11mm}|p{11mm}}
 \hline
 \hline
 & \multicolumn{2}{c||}{CamVid} & \multicolumn{2}{c}{NYUv2} \\
 \hline
 \bf Method &\bf FLOPs &\bf mIoU &\bf FLOPs &\bf mIoU \\
 \hline
 Teacher & 41.0G & 0.594 & 41.0G & 0.517 \\
 KD-ORI & 5.54G & 0.535 & 5.54G & 0.380 \\
 KD-REL & 5.54G & 0.475 & 5.54G & 0.396 \\ 
 KD-UNR & 5.54G & 0.406 & 5.54G & 0.265 \\
 \hline
 RANDOM & 5.54G & 0.018 & 5.54G & 0.021 \\
 DAFL & 5.54G & 0.010 & 5.54G & 0.105 \\
 Ours & 5.54G & \bf 0.535 & 5.54G & \bf 0.364 \\
 \hline
 \end{tabular}
 
 \caption{The mIoU of DeepLabv3 model on CamVid and NYUv2. The teacher model are pretrained on ImageNet when the student are randomly initialized.}\label{seg-miou}
 \vspace{-3mm}
\end{table}

\begin{figure}[t]
 \centering
 \noindent\includegraphics[width=80mm]{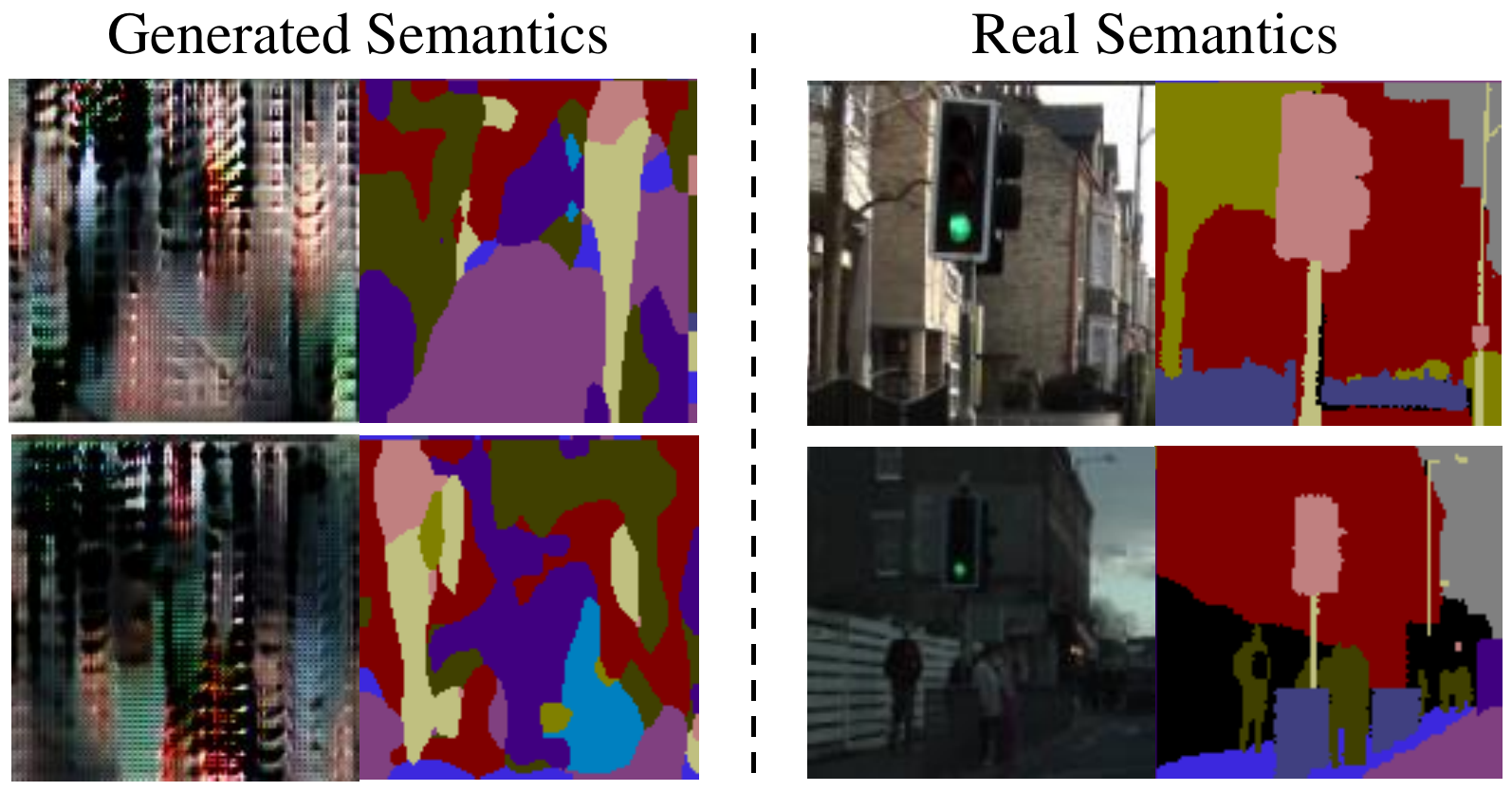}
 \caption{The generated samples on CamVid, as well as their semantics predicted by the teacher model. The co-occurrence of traffic lights and poles, as in real samples, is captured by generator.}
 \label{fig:seg-patterns}
 \vspace{-5mm}
\end{figure}

The main difficulty DAFL encounters is that the one-hot constraint is detrimental to segmentation tasks, in which each pixel has strong correlations with its neighboring ones. In our framework, the generator are encouraged to produce complicated patterns by combining multiple pixels to make the game more challenging. As shown in Figure \ref{fig:seg-patterns}, the generator for CamVid indeed catches the co-occurrence of traffic lights and poles with reasonable spatial correlations. To further study these generated samples, we also conduct an experiment to train a student model with a fixed generator obtained from adversarial distillation and train a student model with mIoU of 0.460. It demonstrates that the generator indeed learns "what should be generated." 

\section{Conclusions}

This paper intoroduces a data-free adversarial distillation framwork for model compression. We propose a novel method to estimate the optimizable upper bound of the intractable model discrepancy between the teacher and the student. Without any access to real data, we successfully reduce the discrepancy by optimizing the upper bound and obtain a comparable student model. Our experiments on classification and segmentation demonstrate that our framework is highly scalable and can be effectively applied to different network architectures. To the best of our knowledge, it is also the first effective data-free method for semantic segmentation. However, it is still very difficult to generate complicated samples. We believe that introducing human priori can effectively improve the generator by avoding useless search space. In the future, we will explore the impact of different prior information on the proposed adversarial distillation framework.

{\small
 \bibliographystyle{ieee_fullname}
 \bibliography{egbib}
}

\clearpage
\appendix

\section*{Supplementary Material}

This supplementary material is organized as follows: 
Sec.~\ref{model_architures} provides model architectures and implementation details for each dataset; Sec.~\ref{hyperparams} presents the influence of different batch sizes on our proposed method. Sec.~\ref{visualization} visualizes more generated samples and segmentation results. 

\section{Model Architectures and Hyperparameters} \label{model_architures}

Table~\ref{arch_summarization} summarizes the basic configurations for each dataset. In our experiments, teacher models are obtained from labeled data, when student models and generators are trained without access to real-world data. We validate our models every 50 iterations. For the sake of simplicity, we regard such a period as an \textbf{``epoch''}. 

\subsection{Generators}

As illustrated in Fig. \ref{generator}, two kinds of vanilla generator architectures are aodpted in our experiments. The first generator, denoted as \textbf{``Generator-A''}, uses Nearest Neighbor Interpolation for upsampling. The second one, denoted as \textbf{``Generator-B''}, is isomorphic to the generator proposed by DCGAN~\cite{radford2015unsupervised}, which replaces the interpolations with deconvolutions. We use the Generator-A for MNIST and CIFAR, and apply the more powerful Generator-B to other datasets. The slope of LeakyReLU is set to 0.2 for more stable gradients. In distillation, all generators are optimized with Adam~\cite{kingma2014adam} with a learning rate of 1e-3. The betas are set to their default values, which are 0.9 and 0.999. 

\subsection{Teachers and Students}

\noindent\textbf{MNIST.} Table \ref{LeNet} provides detailed information about the architectures of LeNet-5 and LeNet-5-Half. In distillation, we use SGD with a fixed learning rate of 0.01 to train the student model for 40 epochs. 

\noindent\textbf{CIFAR10 and CIFAR100.} The modified ResNet~\cite{he2016deep} architectures with 8$\times$ downsampling for CIFAR10 and CIFAR100 are presented in Table \ref{resnet-8x}. The learning rate starts from 0.1 and is divided by 10 at 100 epochs and 200 epochs. We apply a weight decay of 5e-4 and train the student model for 500 epochs.

\def\SPSB#1#2{\rlap{\textsuperscript{#1}}\SB{#2}}
\def\SP#1{\textsuperscript{#1}}
\def\SB#1{\textsubscript{#1}}

\noindent\textbf{Caltech101.} The standard ResNet~\cite{he2016deep} architectures with 32$\times$ downsampling are adopted for Caltech101. During training, the learning rate of SGD is assigned as 0.05 and is decayed every 100 epochs. The student model and generator are optimized for 300 epochs with a weight decay of 5e-4.

\begin{table}[h]
  \centering
  \setlength\extrarowheight{1pt}
  \resizebox{0.95\linewidth}{!}{
  \begin{tabular}{c|c}
  \hline
  \hline
  \bf  \bf Generator-A & \bf Generator-B  \\
  \hline
  FC, Reshape, BN      & FC, Reshape, BN  \\
  Upsample $\uparrow_{2\times}$  & 3$\times$3 512 Deconv $\uparrow_{2\times}$, BN, LReLU\\
  \hline
  3$\times$3 128 Conv, BN, LReLU & 3$\times$3 256 Deconv $\uparrow_{2\times}$, BN, LReLU  \\
  Upsample $\uparrow_{2\times}$    & 3$\times$3 128 Deconv $\uparrow_{2\times}$, BN, LReLU \\
  \hline
  3$\times$3 64 Conv, BN, LReLU  & 3$\times$3 64 Deconv $\uparrow_{2\times}$, BN, LReLU   \\
  3$\times$3 3 Conv, Tanh, BN    & 3$\times$3 3 Conv, Tanh \\ 
  \hline
  \end{tabular}
  }
  \caption{Generator Architectures. The vector input is firstly projected to a feature maps~\cite{radford2015unsupervised} and then upsampled to the required size.}\label{generator}
  \vspace{-0.5mm}
\end{table}

\begin{table}[t]
  \centering
  \setlength\extrarowheight{1pt}
  \resizebox{0.95\linewidth}{!}{
  \begin{tabular}{c|c|c}
  \hline
  \hline
  \bf \bf Output Size & \bf LeNet-5  & \bf LeNet-5-Half \\
  \hline
  \multirow{2}{*}{14$\times$14} & 5$\times$5 6 Conv, ReLU  & 5$\times$5 3 Conv, ReLU \\
  & maxpool $\downarrow_{2\times}$ & maxpool $\downarrow_{2\times}$  \\
  \hline
  \multirow{2}{*}{5$\times$5} & 5$\times$5 16 Conv, ReLU  & 5$\times$5 8 Conv, ReLU  \\
  & maxpool $\downarrow_{2\times}$ & maxpool $\downarrow_{2\times}$  \\
  \hline
  1$\times$1 & 5$\times$5 120 Conv, ReLU & 5$\times$5 60 Conv, ReLU \\
  \hline
  1$\times$1 & 84 FC     &  42 FC        \\
  \hline
  1$\times$1 & \multicolumn{2}{c}{Output FC}    \\
  \hline
  \end{tabular}
  }
  \caption{LeNet-5 architectures for MNIST.}\label{LeNet}
  \vspace{-0.5mm}
\end{table}

\newcommand{\blocka}[2]{\multirow{3}{*}{\(\left[\begin{array}{l}\text{3$\times$3 #1 Conv, BN, ReLU}\\[-.1em] \text{3$\times$3 #1 Conv, BN, ReLU} \end{array}\right]\)$\times$#2}}
  
\renewcommand\arraystretch{1.1}
\setlength{\tabcolsep}{3pt}

\begin{table}[t]
\begin{center}
\resizebox{0.95\linewidth}{!}{
\begin{tabular}{c|c|c}
  \hline
  \hline
  \bf Output Size & \bf ResNet-18-8x & \bf ResNet-34-8x \\
  \hline
  32$\times$32 & \multicolumn{2}{c}{3$\times$3 64 Conv, BN, ReLU}\\
  \hline
  \multirow{3}{*}{32$\times$32} & \blocka{64}{2}{}  & \blocka{64}{3}{} \\
  &  & \\
  &  & \\
  \hline
  \multirow{3}{*}{16$\times$16}  & \blocka{128}{2}  & \blocka{128}{4} \\
  &  &  \\
  &  &  \\
  \hline
  \multirow{3}{*}{8$\times$8}  & \blocka{256}{2}  & \blocka{256}{6} \\
  &  &   \\
  &  &  \\
  \hline
  \multirow{3}{*}{4$\times$4}  & \blocka{512}{2}  & \blocka{512}{3} \\
  &  &  \\
  &  &  \\
  \hline
  1$\times$1 & \multicolumn{2}{c}{Average Pool}  \\
  \hline
  1$\times$1 &  \multicolumn{2}{c}{Output FC} \\
  \hline
\end{tabular}
}
\end{center}
\caption{ResNet architectures with 8$\times$ downsampling for CIFAR10 and CIFAR100.}
\label{resnet-8x}
\vspace{-4mm}
\end{table}

\begin{table*}[t]
  \centering
  \begin{tabular}{c|c|c|c|c|c|c|c|c}
  \hline
  \hline
  \bf Dataset &\bf Teacher &\bf Student &\bf Generator &\bf Input Size &\bf Batch Size &\bf lr\SB{S} & \bf lr\SB{G} & \bf wd \\
  \hline
  MNIST    & LeNet-5 & LeNet-5-Half & Generator-A & $32\times32$ & 512 & 0.01 & \multirow{6}{*}{1e-3} & - \\
  CIFAR10  & ResNet34-8x & ResNet18-8x & Generator-A  & $32\times32$ & 256 & 0.1 & & 5e-4\\
  CIFAR100 & ResNet34-8x & ResNet18-8x & Generator-A  & $32\times32$ & 256 & 0.1 & & 5e-4\\ 
  Caltech101 & ResNet34 & ResNet18 & Generator-B & $128\times128$ & 64 & 0.05 & & 5e-4\\
  CamVid   & DeepLabv3-ResNet50 & DeepLabv3-MobileNetV2 & Generator-B & $256\times341$ & 64 & 0.1 & & 5e-4\\
  NYUv2    & DeepLabv3-ResNet50 & DeepLabv3-MobileNetV2 & Generator-B & $256\times341$ & 64 & 0.1 & & 5e-5\\
  \hline
  \end{tabular}
  \caption{Configurations and hyperparameters for different datasets.}\label{arch_summarization}
  \vspace{-3mm}
\end{table*}

\noindent\textbf{CamVid.} We use DeepLabV3~\cite{chen2017rethinking} with dilated convolutions to tackle this segmentation problem. In distillation, we construct a MobileNet-V2~\cite{sandler2018mobilenetv2} student model and optimize it for 300 epochs using SGD with a learning rate of 0.1 and a weight decay of 5e-4. The learning rate of SGD and Adam is decayed every 100 epochs. 

\noindent\textbf{NYUv2.} We adopt the same architectures and hyperparameters as NYUv2 for NYUv2, except that the learning rate of SGD is modified to 0.05 and the weight decay is reduced to 5e-5. We train the student model and the generator for 300 epochs and multiply the learning rate by 0.3 at 150 epochs and 250 epochs.

\section{Influence of Different Batch Sizes} \label{hyperparams}

\begin{figure}[t]
  \centering
  \noindent\includegraphics[width=8.2cm]{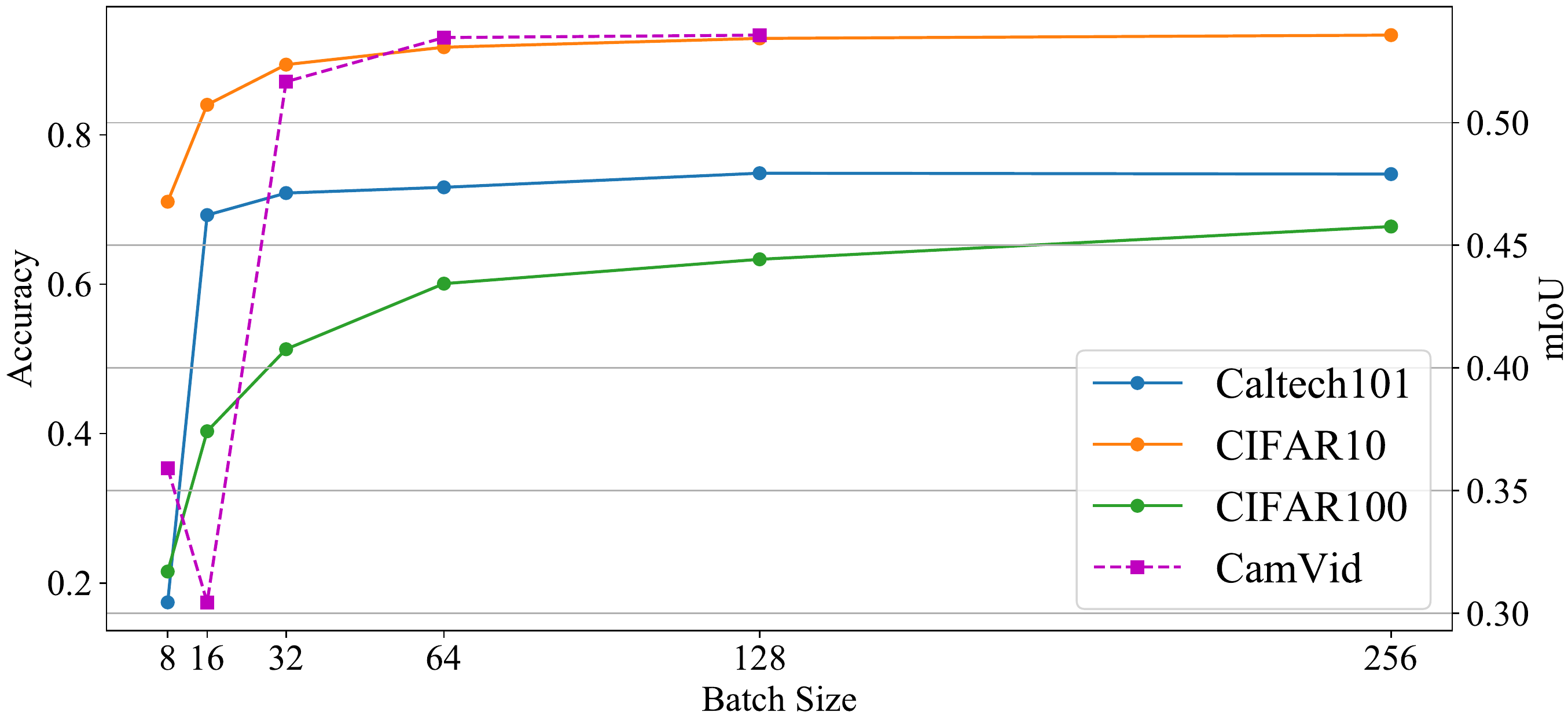}
  \caption{The influence of different batch sizes on our method.}
 \label{batch-size}
 \vspace{-4mm}
\end{figure}

In our method, a large batch size is required to train the generator~\cite{brock2018large} and ensure the accuracy of the discrepancy estimation. To explore the influence of different batch sizes, we conduct several experiments on classification and semantic segmentation datasets. As illustrated in Fig.~\ref{batch-size}, a small batch size injures the performance of student models. We also found that increasing batch size can bring tremendous benefits to our method. An important reason for the phenomenon is that the large batch size can provide sufficient statistical information for hard sample generation and make the training more stable.

\section{More Visualization} \label{visualization}

\begin{figure*}[t]
  \centering
  \noindent\includegraphics[width=14cm]{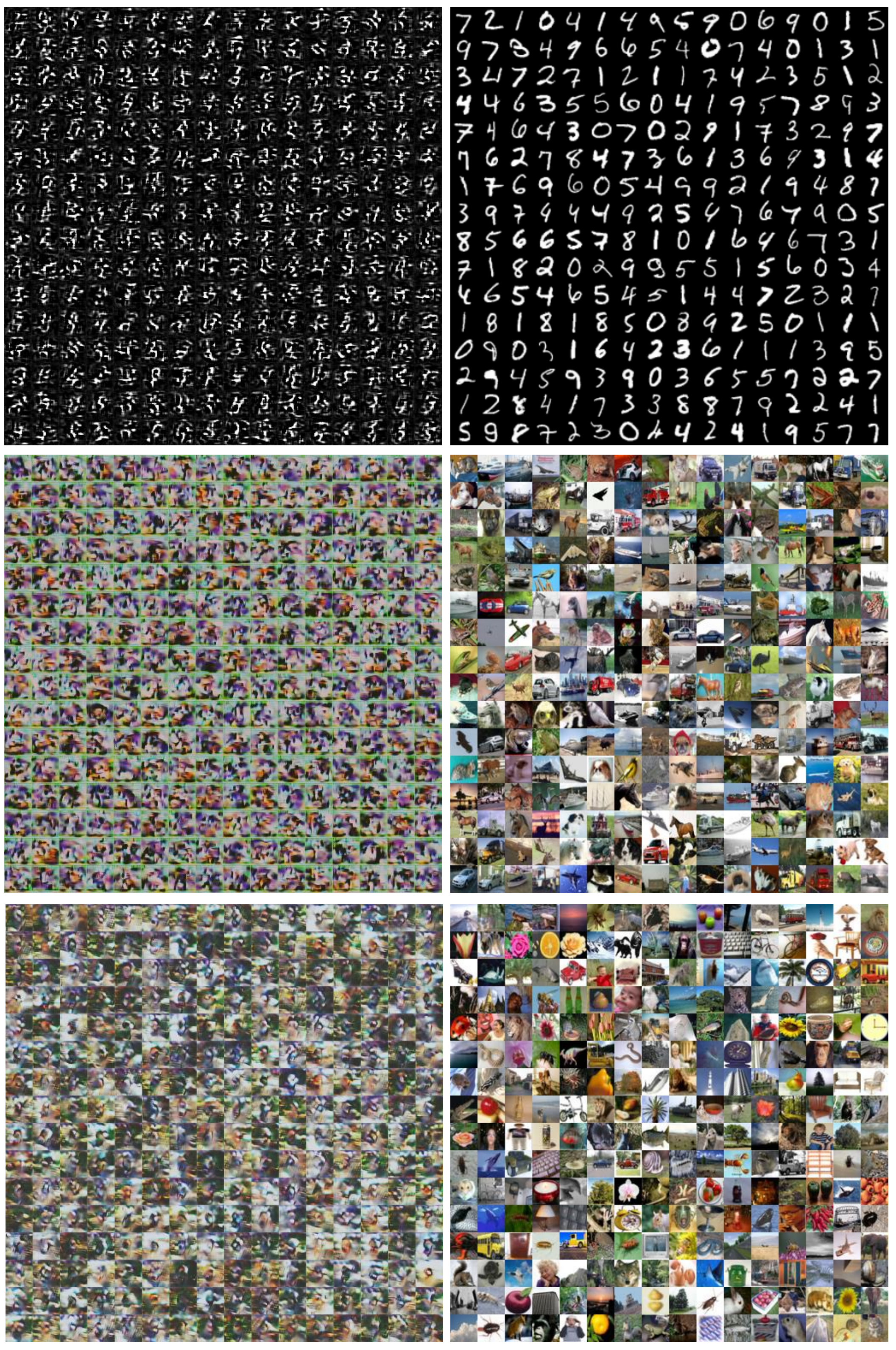}
  \caption{Generated samples (left) and real samples (right) from MNIST, CIFAR10 and CIFAR100. }
 \label{vis-cls}
 \vspace{-3mm}
\end{figure*}

\begin{figure*}[t]
  \centering
  \noindent\includegraphics[width=14cm]{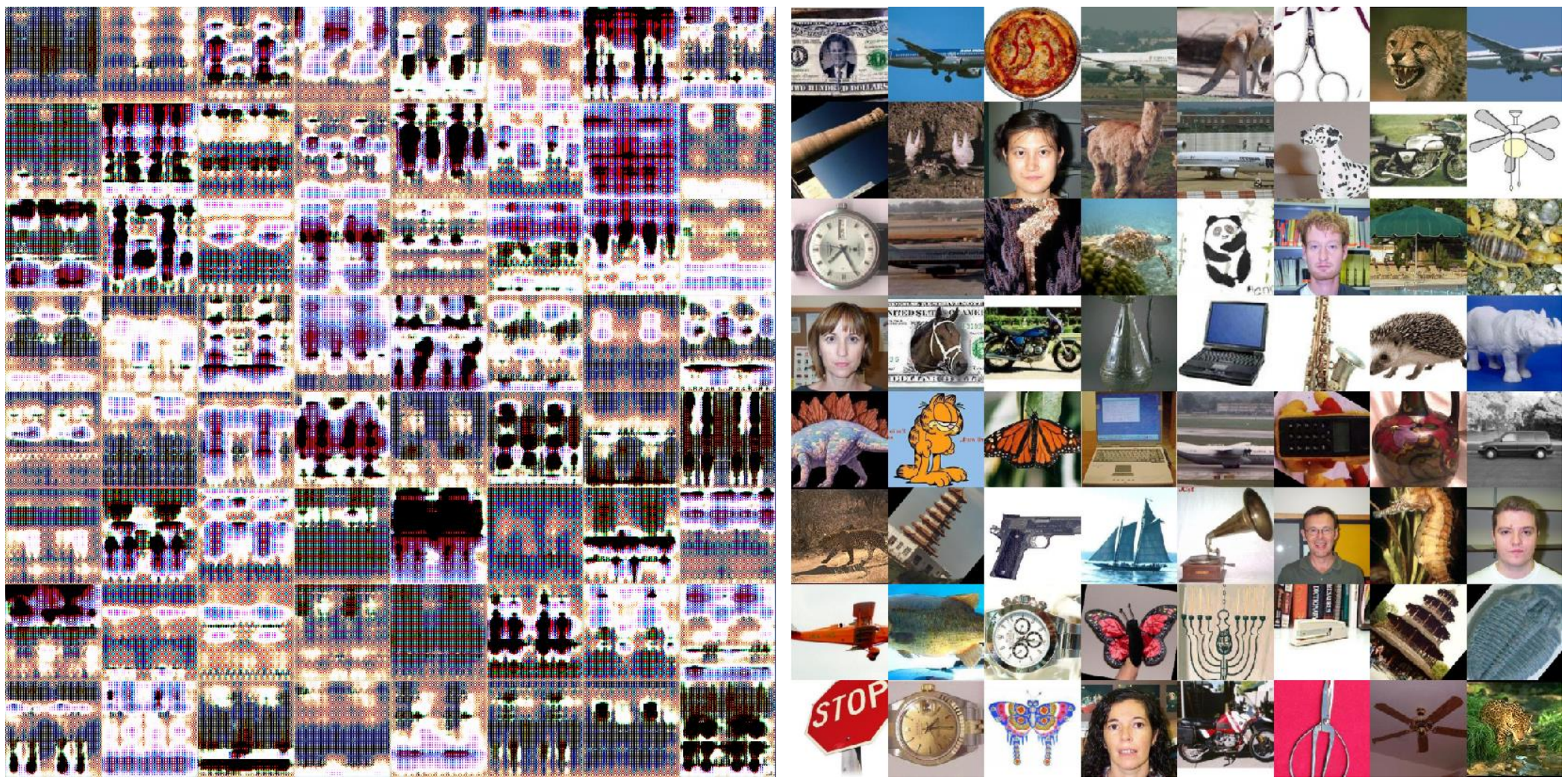}
  \caption{Generated samples (left) and real samples (right) from Caltech101.}
 \label{caltech101-vis}
 \vspace{-3mm}
\end{figure*}

\begin{figure*}[t]
  \centering
  \noindent\includegraphics[width=15cm]{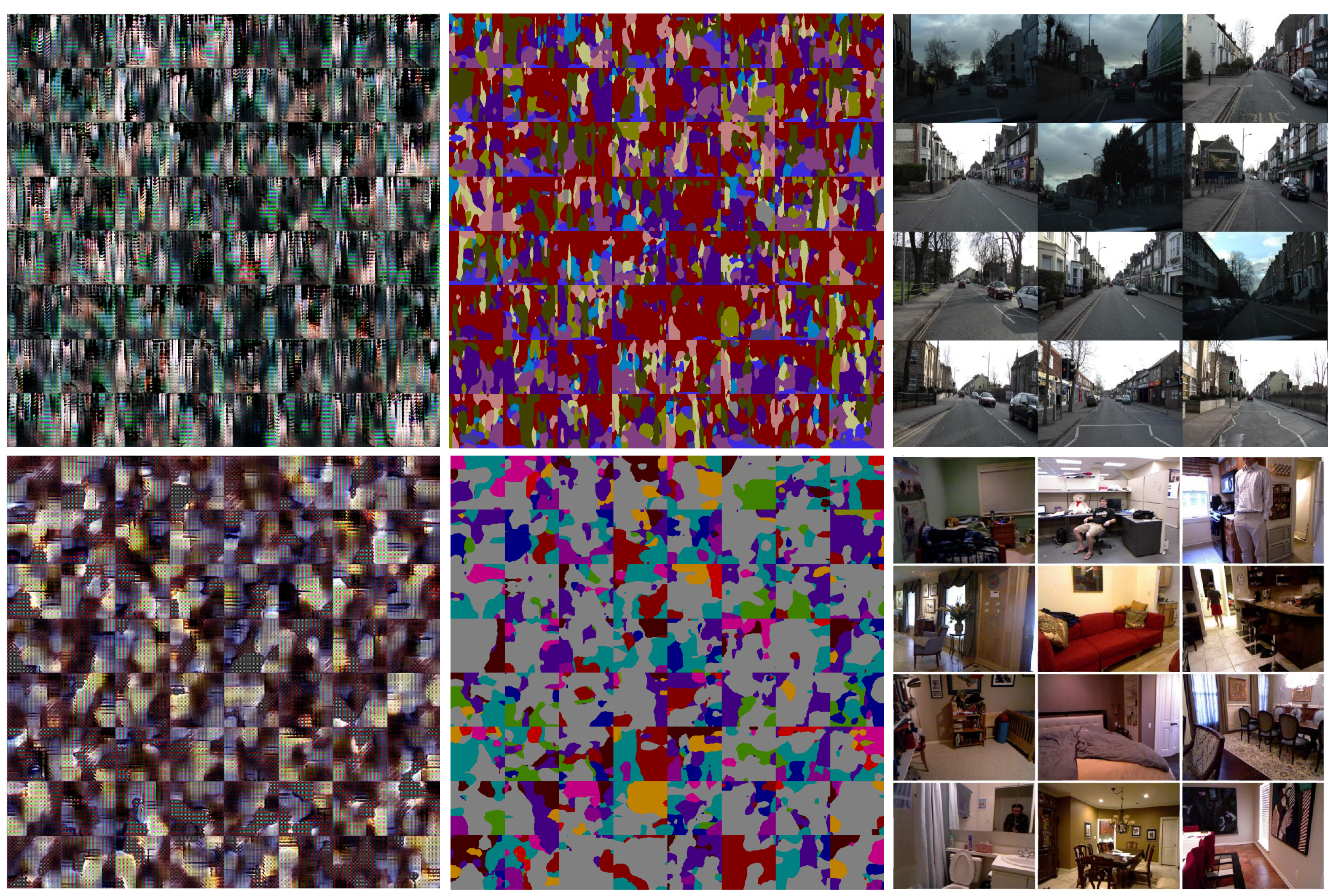}
  \caption{Generated samples (left), teacher predictions (middle) and real samples (right) from CamVid and NYUv2.}
 \label{vis-seg}
 \vspace{-3mm}
\end{figure*}

\begin{figure*}[t]
  \centering
  \noindent\includegraphics[width=15cm]{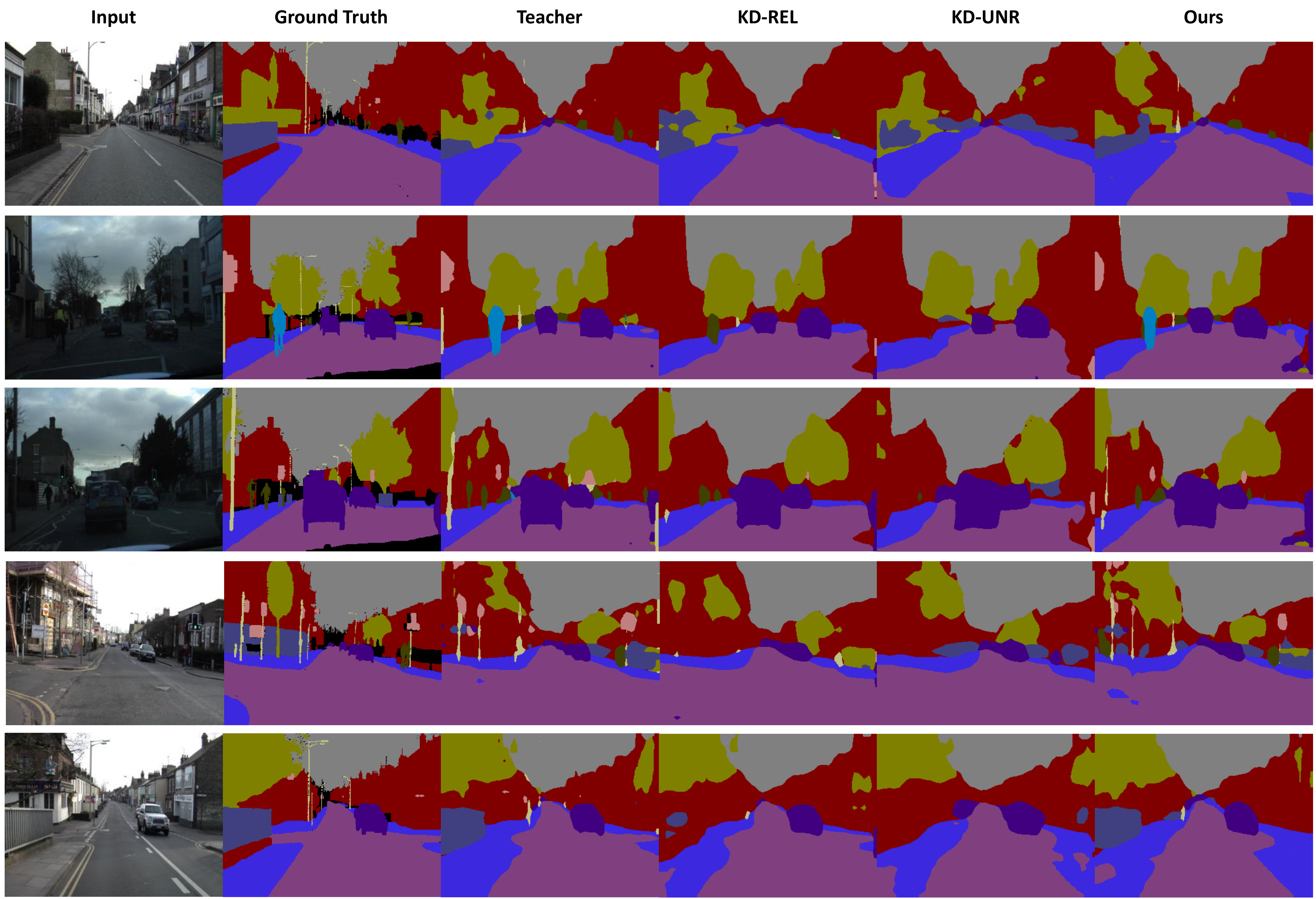}
  \caption{Segmentation results on CamVid. KD-REL uses Cityscapes as  training data and KD-UNR adopts VOC2012 as an alternative.}
 \label{seg-camvid}
 \vspace{-3mm}
\end{figure*}

\begin{figure*}[t]
  \centering
  \noindent\includegraphics[width=15cm]{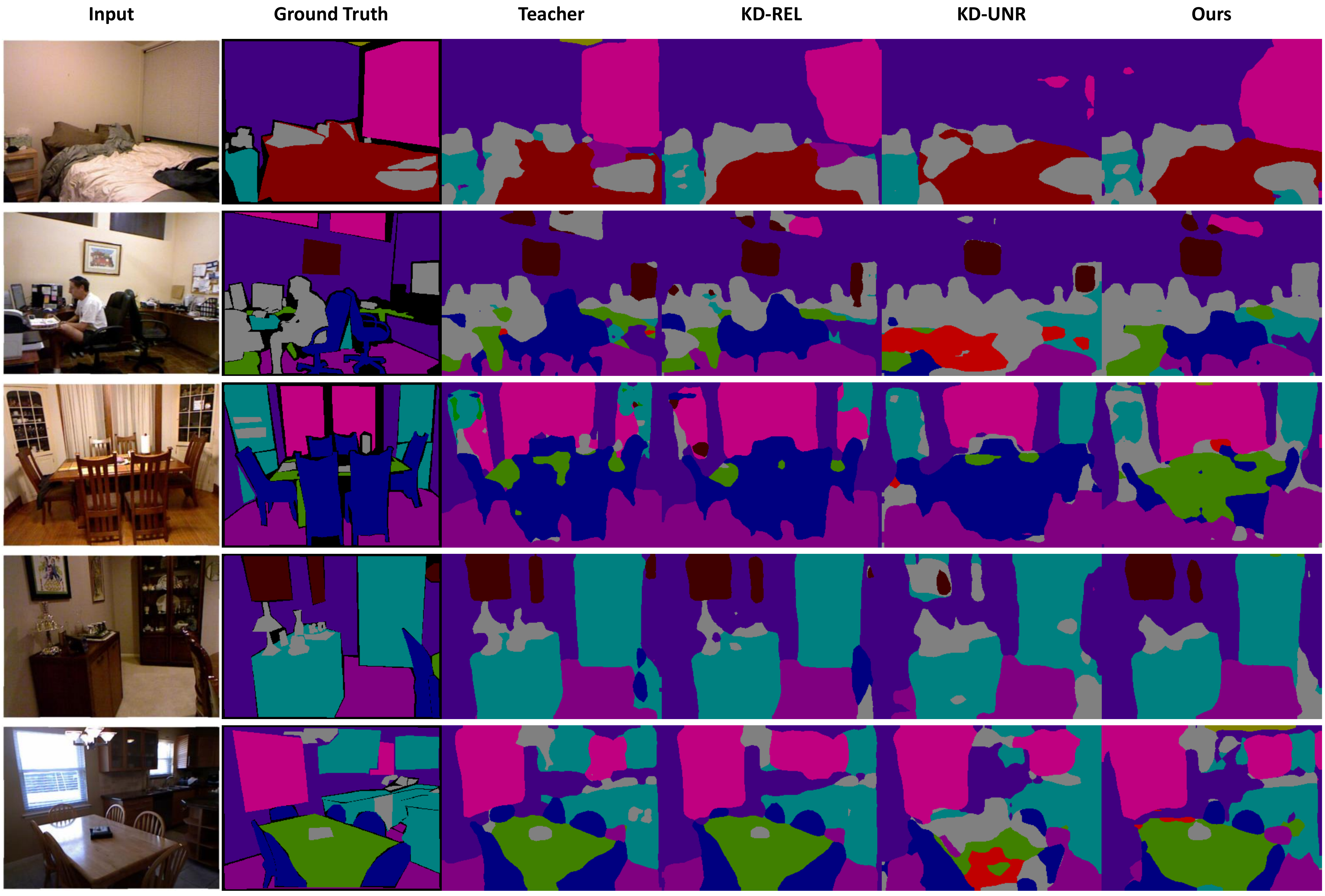}
  \caption{Segmentation results on NYUv2. KD-REL uses SunRGBD as training data and KD-UNR adopts VOC2012 as an alternative.}
 \label{seg-nyuv2}
 \vspace{-3mm}
\end{figure*}

We provide more visualization results in this section. Fig.~\ref{vis-cls} and Fig.~\ref{caltech101-vis} compare the generated samples with real samples on classification datasets. Those fake samples can not be recognized by humans, but indeed contain sufficient knowledge for their tasks. Fig.~\ref{vis-seg} provides some generated samples on segmentation datasets, as well as the predictions produced by teacher models.
To further demonstrate the effectiveness of our method, we provide more segmentation results in Fig.~\ref{seg-camvid} and Fig.~\ref{seg-nyuv2}. As in the main paper, our data-free method is compared with several data-driven baselines, such as KD-REL and KD-UNR. KD-REL requires related data and KD-UNR uses some unrelated data. 

\end{document}